\documentclass[lettersize,journal]{IEEEtran}
\usepackage{amsmath,amsfonts}
\usepackage{algorithmic}
\usepackage{algorithm}
\usepackage{array}
\usepackage[caption=false]{subfig}
\usepackage{textcomp}
\usepackage{stfloats}
\usepackage{url}
\usepackage{verbatim}
\usepackage{graphicx}
\usepackage{cite}
\usepackage[table]{xcolor}
\usepackage{multirow}
\usepackage[pagebackref,breaklinks,colorlinks]{hyperref}

\definecolor{greencell}{RGB}{204, 255, 153}
\definecolor{redcell}{RGB}{255, 153, 153}

\newcommand{\dataset}{WASD}
\newcommand{\model}{BIAS}

\begin{document}

\title{BIAS: A Body-based Interpretable Active Speaker Approach}

\author{Tiago Roxo,~Joana C. Costa,~Pedro R. M. Inácio,~\IEEEmembership{Senior Member,~IEEE,}~Hugo Proença,~\IEEEmembership{Senior Member,~IEEE}\\
Instituto de Telecomunicações, University of Beira Interior, Portugal\\
{\tt\small \{tiago.roxo, joana.cabral.costa\}@ubi.pt, \{inacio, hugomcp\}@di.ubi.pt}
\thanks{Manuscript received XX, 2024; revised XX, 2024.}
}


\markboth{IEEE Transactions on Biometrics, Behavior, and Identity Science,~Vol.~XX, No.~XX, XX~2024}%
{Roxo \MakeLowercase{\textit{et al.}}: BIAS: A Body-based Interpretable Active Speaker Approach}


\maketitle


\begin{abstract}

State-of-the-art Active Speaker Detection (ASD) approaches heavily rely on audio and facial features to perform, which is not a sustainable approach in \textit{wild} scenarios. Although these methods achieve good results in the \emph{standard} AVA-ActiveSpeaker set, a recent wilder ASD dataset (WASD) showed the limitations of such models and raised the need for new approaches. As such, we propose \model{}, a model that, for the first time, combines audio, face, and body information, to accurately predict active speakers in varying/challenging conditions. Additionally, we design \model{} to provide interpretability by proposing a novel use for Squeeze-and-Excitation blocks, namely in attention heatmaps creation and feature importance assessment. For a full interpretability setup, we annotate an ASD-related actions dataset (ASD-Text) to finetune a ViT-GPT2 for text scene description to complement \model{}~interpretability. The results show that \model{}~is state-of-the-art in challenging conditions where body-based features are of utmost importance (Columbia, open-settings, and WASD), and yields competitive results in AVA-ActiveSpeaker, where face is more influential than body for ASD. \model{}~interpretability also shows the features/aspects more relevant towards ASD prediction in varying settings, making it a strong baseline for further developments in interpretable ASD models, and is available at \url{https://github.com/Tiago-Roxo/BIAS}.

\end{abstract}



\section{Introduction}
\label{sec:intro}

\IEEEPARstart{C}{urrent} Active Speaker Detection (ASD) models are known to  perform reliably using only audio and face-based information. This is mainly due to the fact that state-of-the-art ASD datasets have good audio and face quality, yielding from controlled setups (movies, from AVA-ActiveSpeaker~\cite{roth2020ava}) and cooperative settings (interviews, from AWS~\cite{kim2021look}). Recently, WASD~\cite{wasd} has been announced as a more challenging set, with degraded audio and face data quality, corresponding to less constrained data acquisition scenarios.

\begin{figure}[t]
  \centering
  \includegraphics[width=0.9\linewidth]{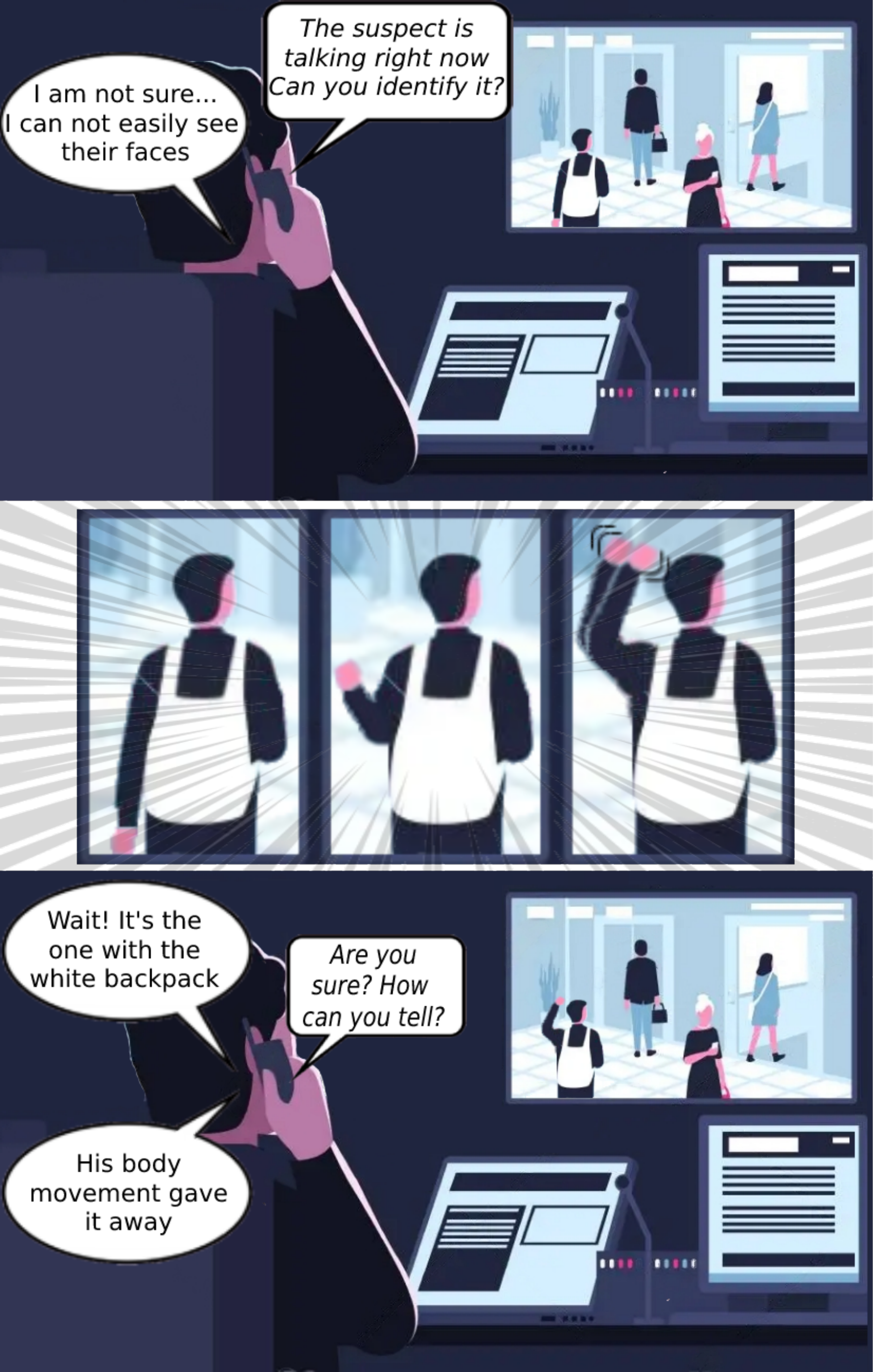}

   \caption{Illustration of the \textbf{BIAS insight}: in surveillance settings, where facial and audio-based features might not be always available, body data should be  crucial to accurately detect the active speakers. In such challenging conditions, providing reliable explanations for the reasoning behind the provided responses is also an important feature. This paper describes \model{}, which singularly uses facial, audio, and body-based features, also providing visual interpretability and feature importance assessment for its responses.}
   \label{fig:init_image}
\end{figure}

In this context, the existing models are not suitable for wilder settings, where audio quality might be poor and faces occluded (Figure~\ref{fig:init_image}). As such, we propose BIAS, an approach for ASD, that uses, for the first time, body data to complement face and audio-based features, achieving state-of-the-art results in challenging sets (WASD and open-settings of Columbia), and competitive results in more cooperative conditions (AVA-ActiveSpeaker), where the body relevance is reduced since the face is the predominant feature in this data. Furthermore, we propose a novel use of Squeeze-and-Excitation (SE) blocks~\cite{hu2018squeeze} to provide reasoning for model decision and analyze the importance of different features. This way, we are able to obtain visual interpretability and assess feature influence in varying ASD settings, via SE vector manipulation. 
Finally, to improve the interpretability of ASD-related scenarios, we complement the attention of BIAS (visual) with descriptions of existing actions, via caption generation (textual). Given the absence of training data of subtle actions related to ASD tasks (\textit{e.g.}, raise hand, cross arms, open mouth to talk), we use WASD data to annotate an ASD-related actions dataset (ASD-Text) and  finetune a Generative Pretrained Transformer (GPT) model on it which, in conjunction with the visual interpretability of BIAS, can be used for a full interpretability setup. To summarize, the main contributions are:

\begin{itemize}
    \item We propose \model{}, an ASD model based on a long temporal context approach, using audio, face, and body-based information. With respect to the state-of-the-art, \model{}~is innovative by being the first method to use body-based information for ASD, yielding competitive results in AVA-ActiveSpeaker and is state-of-the-art in WASD and Columbia (open settings version);
    \item We propose a novel use of SE blocks for attention heatmap creation (visual interpretability) and perceive feature importance in the decisions, which can be easily included in common and customized architecture, for image and video settings;
    \item To improve the interpretability of ASD-related scenarios, we annotate an ASD-related actions dataset (ASD-Text) for text scene captions, providing a training setup for fine scene description in the ASD context.
\end{itemize}

\section{Related Work}
\label{sec:related-work}

\textbf{ASD Context.}
Given the task to determine the talking speaker from a set of admissible candidates, various state-of-the-art ASD datasets have been recently proposed~\cite{chakravarty2016cross, alcazar2021maas, donley2021easycom, roth2020ava, kim2021look}. Columbia~\cite{chakravarty2016cross} contains 87 minutes of a panel discussion and Talkies~\cite{alcazar2021maas} focuses on low duration videos, totalling 4 hours, with off-screen speaking. Easycom~\cite{donley2021easycom} is designed for multiple augmented reality tasks, composed of various speaking sessions with background noise. AVA-ActiveSpeaker~\cite{roth2020ava} contains Hollywood videos totalling almost 38 hours, with demographic diversity and Frames Per Second (FPS) variation, and has application in other topics such as audio anomaly detection~\cite{roxo2023exploring}. ASW~\cite{kim2021look} has over 30 hours, with videos randomly selected from the VoxConverse~\cite{Chung2020}, containing various sets of interviews. Recently, \dataset~\cite{wasd} has been announced, with 30 hours of data grouped based on the audio and face quality, with balanced demographic diversity and body annotations data. For a broader overview of the ASD context, Robi \textit{et al.}~\cite{robi2024active} review the main ASD modalities, applications and challenges.

\textbf{ASD Models.}
Based on the available data for ASD, current state-of-the-art heavily rely on face and audio data, with audiovisual combination using 3D architectures~\cite{chung2019naver}, hybrid 2D-3D models~\cite{zhang2019multi}, large-scale pretraining~\cite{chung2016out, chung2019perfect}, feature embedding improvement~\cite{hadsell2006dimensionality}, and attention mechanisms~\cite{vaswani2017attention, afouras2020self, cheng2020look}. Various ASD works~\cite{alcazar2020active, kopuklu2021design, zhang2021unicon, alcazar2021maas} based their model on a two-step process, where the first focuses on short-term analysis (audio with face combination) and the second on multi-speaker analysis. ASC~\cite{alcazar2020active} focuses on multi speaker analysis via temporal refinement, ASDNet~\cite{kopuklu2021design} uses a similar approach for inter-speaker relations with improved visual backbones, and UniCon~\cite{zhang2021unicon} relies on audio-visual relational contexts with various backbones. 
The improvement of ASD performance by assessing contextual information via speaker relation using Graph Convolutional Networks (GCN)~\cite{welling2016semi} has also been explored~\cite{alcazar2021maas, min2022learning,alcazar2022end}. Diverging from two-step training, end-to-end models have also emerged for ASD~\cite{tao2021someone, alcazar2022end, min2022learning, liao2023light}. TalkNet~\cite{tao2021someone} focused on improving long-term temporal context with audio-visual synchronization,  EASEE~\cite{alcazar2022end} included GCN to complement spatial and temporal speaker relations, and Light-ASD~\cite{liao2023light} proposed a lightweight model by splitting 2D and 3D convolutions for audio-visual feature extraction, and applied Bidirectional Gated Recurrent Units (BGRU) for cross-modal modeling.

\textbf{Body Information for Attribute Recognition.}
Although recent works on ASD do not use body information, this data contains information that could contribute to improve model performance, particularly in wilder conditions (\textit{e.g.}, surveillance settings), where face is not reliably accessed. Pedestrian Attribute Recognition (PAR) datasets~\cite{deng2014pedestrian,liu2017hydraplus,li2016richly} are examples of these scenarios, containing person cropped images from surveillance settings, used to identify attributes (\textit{e.g.}, clothing, accessory usage, gender, age) under challenging covariates such as occlusion, pose, image resolution, and luminosity. Works in this area focused on different strategies ranging from different architecture combination~\cite{roxo2022yinyang, zhao2018grouping, tang2019improving}, attention-based approaches~\cite{sarafianos2018deep, guo2019visual}, and attribute relation importance~\cite{li2015multi, lin2019improving, jia2020rethinking}.

\textbf{Model Interpretability.} For visual interpretability, we can group methods into two main categories: \textit{gradient based}~\cite{selvaraju2017grad, shrikumar2016not, srinivas2019full, smilkov2017smoothgrad, sundararajan2017axiomatic} (gradients of each layer, computed through backpropagation) and \textit{attribution propagation}~\cite{bach2015pixel, gu2019understanding, iwana2019explaining, lundberg2017unified, shrikumar2017learning} (recursive decomposition of layers contributions, all the way to model's input). Saliency based methods~\cite{zhou2018interpreting, dabkowski2017real}, Excitation Backprop~\cite{zhang2018top}, and Perturbation methods~\cite{fong2019understanding, fong2017interpretable} are also visual interpretable approaches in computer vision, with Transformer-based interpretability~\cite{chefer2021transformer} being recently explored. Although most works explore model interpretability in object classification datasets, its use in face~\cite{yin2019towards, winter2022demystifying}, body~\cite{fu2021learning}, and PAR~\cite{doshi2023towards, roxo2021wildgender} data is not unprecedented. Contrary to current approaches, we propose a SE block-based visual interpretation, obtainable in inference time, without requiring additional computational cost for attention heatmaps creation.


\begin{figure*}[!tb]
\centering
\includegraphics[width=0.99\textwidth]{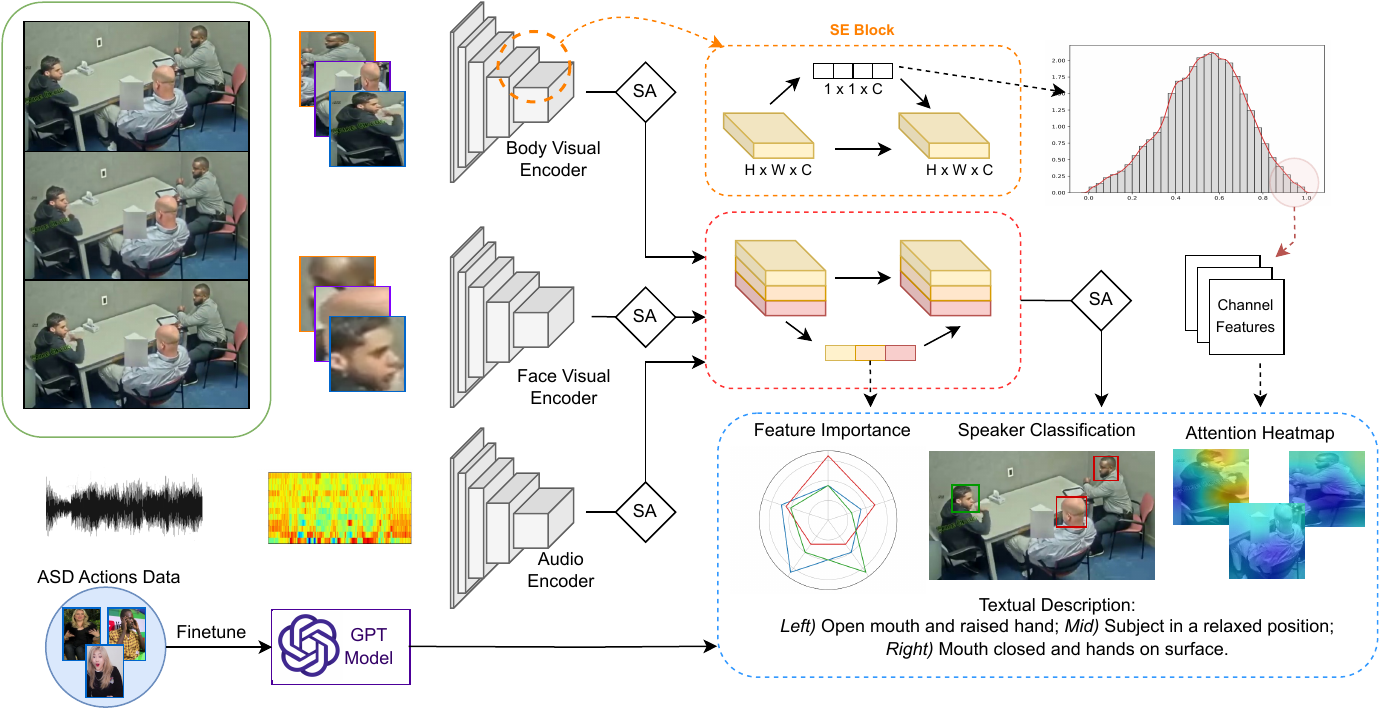}
\caption{
Overview of the \model{}~architecture and pipeline, with GPT model integration: body and face-based data is fed into the respective visual encoders, while audio is processed into MFCC before encoding. SE blocks are used in visual encoders and feature combination for attention heatmaps and feature relative importance, respectively. SA refers to self-attention blocks. Heatmaps are created by combining channel features of the respective top 10\% SE vector values. \model{}~prediction is based on feature combination, accompanied with visual interpretability and feature importance assessment, complemented by text descriptions from a GPT model finetuned in ASD-related actions data (ASD-Text).
}
\label{fig:main_image}
\end{figure*}

\section{\model{}~Approach}
\label{sec:proposed_model}

We propose \model{}, an interpretable model for ASD, based on a long temporal context approach, using audio, face, and body data. We process data with modified encoders and customized feature vector combination, using SE blocks to provide visual interpretability and feature importance assessment, respectively. \model{} is distinctive from other state-of-the-art models by using, for the first time, body data in ASD tasks, translating into state-of-the-art performance in challenging sets and competitive results in cooperative conditions. The overall architecture is displayed in Figure~\ref{fig:main_image}, with details of each part in the following subsections. Note that, although body data also contain facial cues, these are not as easily perceived as body movements (given the face/body proportion), which creates a stronger focus on hands and arms movements.

\subsection{Visual and Audio Encoders}

\textbf{Backbone.} 
We obtain a sequence of frame-based embedding using a customized backbone, based on the ResNet18 architecture, composed of a 3D convolutional layer followed by ResNet18~\cite{he2016deep} layers and a SE block, before average pooling. The inclusion of a SE block is to provide a way to retrieve informative features, 
via assessment of SE vector values, useful for visual interpretability, which we discuss in Section~\ref{sec:interpretability}.

\textbf{Temporal Enconding.} 
For long-term visual spatio-temporal representation, we use a visual temporal network, composed of 5 depth-wise separable convolutional layers (DS Conv1D~\cite{afouras2018deep}) with residual connections, followed by a Conv1D layer for feature dimension reduction (to 128-feature embedding).

\textbf{Audio Encoder.}
We obtain audio temporal encoding using a ResNet34 with SE blocks in its layers (SE-ResNet34). Audio frames are represented as 13-dimensional Mel-Frequency Cepstral Coefficients (MFCCs), with the audio encoder outputting them as a 128-dimensional audio embedding for subsequent visual and audio conjunction.

\subsection{Squeeze-and-Excitation for Interpretability}
\label{sec:interpretability}

The novel use of SE block in \model{}~is included in two parts of its architecture: end of visual backbones (visual interpretability) and feature combination (perceived feature importance).

\begin{figure}[!tb]
     \centering
     \includegraphics[width=.47\textwidth]{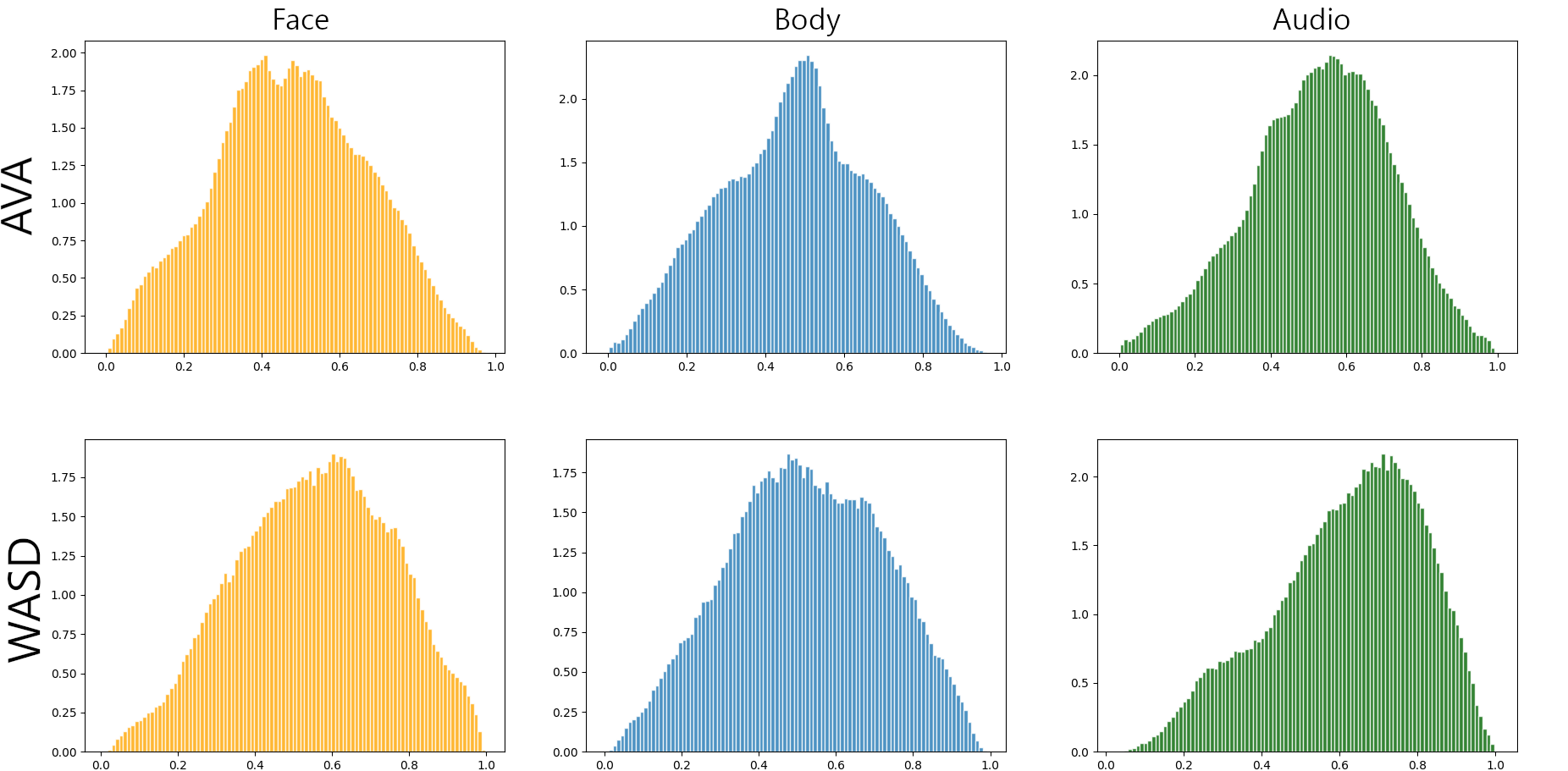}
     \caption{SE vector values from feature (audio, body, and face) combination, for \model{} trained in AVA-ActiveSpeaker and WASD. Both datasets follow a normal distribution. AVA refers to AVA-ActiveSpeaker.}
     \label{fig:normal_distributions}
\end{figure}

\textbf{Visual Interpretability}. In our model, we use SE in encoders to obtain the features (channels) perceived as highly influential towards ASD (\textit{i.e.}, high value in SE vector). However, the perception of \textit{high} is relative to the video/categories used, so we can not use a default/hard-coded value to define what is high or low importance. As such, given that the values of the SE vector follow a Normal distribution (Figure~\ref{fig:normal_distributions}), we use its formula to obtain the top 10\% values of the SE vector (and, consequently, the top 10\% channels) to use for ASD evaluation. The reason for considering the top 10\% in our experiments is to balance model performance and visual interpretability (higher \% leads to worse performance, and lower \% leads to less clear visual interpretability). Finally, we conjugate the selected channels by bicubic interpolation on top of the original image to provide a heatmap of the most important regions for ASD.

\textbf{Perceived Importance and Feature Influence}. Another use of SE in BIAS is to combine audio, face, and body information (embeddings of 128 dimensions). For feature importance assessment, we create three different graphs that display the importance of each feature (audio, face, and body) for the categories of WASD (Table~\ref{table:category-features-dataset}). 
For each feature, we normalize the SE values for visualization purposes to better display the feature importance on WASD categories. The normalization is done based on the values obtained by the model for a video and it is not included in the BIAS training process, meaning that BIAS does not rely on additional/external information and can be used for other domains without requiring adaptations. Regarding feature influence in model performance, when the experiments do not contain a certain feature it means that we set the SE vector values to 0 for the channels of the considered feature (\textit{e.g.}, when we consider \textit{only face}, we set the values to 0 for audio and body channels).

\subsection{Self-attention and Loss Function}

\textbf{Self-attention.} To improve feature importance assessment, we use an attention layer based on query, key, value strategy~\cite{vaswani2017attention}, after visual encoders and feature conjugation. Initially used in the Transformer architecture, this approach has been extensively used for attention purposes, with a role in \model{}~of improving audio-visual correlation. This translates into more accurate distinction between speaking and non-speaking frames, given each speaker.

\textbf{Loss Function.} The main source for ASD is the conjunction of 3 data sources: audio, face, and body. To improve independent feature relevance for ASD, and to not  motivate ASD prediction solely based on their conjunction, we include weighted cross entropy losses using individual data on ASD training prediction. For inference, only the conjunction of 3 data sources is considered.

\subsection{ASD-Text Dataset}
\label{sec:asd-textual}

For text scene description, we finetune a ViT-GPT2~\cite{wu2020visual,radford2019language} model on WASD data, with ViT trained in ImageNet~\cite{imagenet} and GPT2 in WebText~\cite{radford2019language}. Before annotations, we define a set of 14 admissible ASD-related actions, in Table~\ref{table:action-gpt}. Then, for each action we create 3 admissible captions for each gender to ease data annotation, and construct a Graphical User Interface (GUI) to annotate. For each subject image, the annotators select the gender and all existing actions on it, using the GUI to create annotations with predefined admissible captions from the selected actions and gender. The total annotated data is composed of 47 246 captions, and 11 733 images from WASD subvideos (up to 30s), randomly selected from all videos of the WASD train set. Similar to the COCO captions dataset~\cite{chen2015microsoft}, we divide our annotations into 90/10 train/test~\cite{karpathy2015deep}. The caption prediction of ViT-GPT2 model in Section~\ref{sec:text-description} is done on images from the test set of WASD.

\begin{table}[!tb]
    \centering
    \small
    \renewcommand{\arraystretch}{1.05}
    \caption{ASD-Text dataset action labels, grouped by body parts.}
    \begin{tabular}{c*{3}{c}}\hline 

      \textbf{Body Part} &
      \textbf{\#} &
      \textbf{Action Label} &
       \\
        \hline\hline

        \multirow{4}{*}{Hand} &
        1 & 
        Hand raised in the air \\
        
        & 2 & 
        Hand touching the face \\
        
        & 3 & 
        Hand raised with object \\
        
        & 4 & 
        Hand movement (not raised) \\
        \hline

        \multirow{5}{*}{Mouth} &
        5 & 
        Mouth occlusion with object \\
        
        & 6 & 
        Mouth occlusion with hand \\
        
        & 7 & 
        Mouth move from speech \\
        
        & 8 & 
        Mouth move from expression \\
        
        & 9 & 
        Mouth not moving \\
        \hline
        
        \multirow{2}{*}{Arms} &
        10 & 
        Crossed arms \\

        & 11 & 
        Arms behind back \\
        \hline

        \multirow{3}{*}{Body} &
        12 & 
        Body in relaxed position  \\

        & 13 & 
        Body facing forward  \\

        & 14 & 
        Wild body movement  \\
    
        \hline
        
    \end{tabular}
    \label{table:action-gpt}
\end{table}

\subsection{Implementation Details}

\model{} is trained with an Adam optimizer, with a initial learning rate of 10$^{-4}$, decreasing 5\% for each epoch. All visual data is reshaped into 112 x 112, audio data is represented by 13-dimensional MFCC, and both visual and audio features have an encoding dimension of 128. Self-attention uses a transformer layer with 8 attention heads. For visual augmentation, we perform random flip, rotate and crop, while for audio augmentation, we use negative audio sampling~\cite{tao2021someone}. In sum, given a video data during training, a audio track of a new one is randomly selected from the same batch as noise, maintaining the same speaking label of the original soundtrack. Since AVA-ActiveSpeaker does not have body data annotations, we obtain body bounding box annotations from AVA Actions Dataset~\cite{gu2018ava} and complement them with speaking labels of AVA-ActiveSpeaker. ViT-GP2 model is finetuned for 3 epochs, using AdamW optimizer, with a learning rate of 5$\times$10$^{-5}$, without weight decay or warmup steps.

\begin{table}[!tb]
    \centering
    \small
    \renewcommand{\arraystretch}{1.05}
    \caption{Category feature matrix. Feature description: FA, Face Availability; SO, Speech Overlap;  DS, Delayed Speech; FO, Facial Occlusion; HV, Human Voice as Background Noise; SS, Surveillance Settings. The absence of a certain feature is presented with $\times$, while its presence with $\checkmark$. Features containing $\mathord{?}$ refer to non-guarantee of its presence or absence. Green cells refer to features favorable for ASD, while red ones are unfavorable. Retrieved from \cite{wasd}.}
    \begin{tabular}{c*{6}{c}}\hline
    
        \makebox[4em]{\textbf{Category}} &
        FA & SO	& DS & FO & HV & SS  \\
        \hline\hline
        
        Optimal Conditions (OC) & 
        \cellcolor{greencell}$\checkmark$ & \cellcolor{greencell}$\times$ & \cellcolor{greencell}$\times$ & \cellcolor{greencell}$\times$ & \cellcolor{greencell}$\times$ & \cellcolor{greencell}$\times$  \\
        
        Speech Impairment (SI) & 
        \cellcolor{greencell}$\checkmark$ & \cellcolor{redcell}$\checkmark$ & \cellcolor{redcell}$\checkmark$ & \cellcolor{greencell}$\times$ & \cellcolor{greencell}$\times$ & \cellcolor{greencell}$\times$  \\
        
        Face Occlusion (FO) & 
        \cellcolor{greencell}$\checkmark$ & \cellcolor{greencell}$\times$ & \cellcolor{greencell}$\times$  & \cellcolor{redcell}$\checkmark$ & \cellcolor{greencell}$\times$ &  \cellcolor{greencell}$\times$  \\
        
        Human Voice Noise (HVN) & 
        \cellcolor{greencell}$\checkmark$ & \cellcolor{greencell}$\times$ & \cellcolor{greencell}$\times$ & \cellcolor{greencell}$\times$ & \cellcolor{redcell}$\checkmark$ &  \cellcolor{greencell}$\times$  \\

        Surveillance Settings (SS) &
        \cellcolor{redcell}$\mathord{?}$ & \cellcolor{redcell}$\mathord{?}$ & \cellcolor{redcell}$\mathord{?}$ & \cellcolor{redcell}$\mathord{?}$ & \cellcolor{redcell}$\mathord{?}$ &\cellcolor{redcell} $\checkmark$  \\
    
        \hline
        
    \end{tabular}
    \label{table:category-features-dataset}
\end{table}

\section{Experiments}
\label{sec:experiments}

\subsection{Datasets, Models, and Evaluation Metrics}
\label{dataset}

\textbf{Datasets.} The AVA-ActiveSpeaker dataset~\cite{roth2020ava} is an audio-visual active speaker dataset from Hollywood movies. With 262 15 minute videos, typically only train and validation sets are used for experiments: 120 for training, and 33 for validation, corresponding to 29,723 and 8,015 video utterances, respectively, ranging from 1 to 10 seconds. 
The main challenges of this dataset are related to language diversity, FPS variation, the existence of faces with low pixel numbers, blurry images, noisy audio, and dubbed dialogues. Similar to other works, we report the obtained results on the AVA-ActiveSpeaker validation subset. 

The \dataset~dataset~\cite{wasd} compiles a set of videos from real interactions with varying accessibility of the two components for ASD: \textit{audio} and \textit{face}. With 30 hours of labelled data, \dataset~is divided into 5 categories with varying degrees of audio and face quality, grouped into categories: Optimal Conditions (OC), Speech Impairment (SI), Face Occlusion (FO), Human Voice Noise (HVN), and Surveillance Settings (SS). Table~\ref{table:category-features-dataset} presents the main characteristics of \dataset~categories. \dataset~contains 164 videos, with varying FPS, averaging 28 second duration, with balanced demographics, and similar train/test division as AVA-ActiveSpeaker (80/20). We report the results on each category and Easy-Hard grouping, following \dataset~experiments (Easy: OC and SI, Hard: FO, HVN, and SS).

We also consider Columbia~\cite{chakravarty2016cross} following the methodology of Light-ASD~\cite{liao2023light} where models are trained in AVA-ActiveSpeaker, without any additional fine-tuning. Columbia consists of an 87-minute panel discussion video, with five speakers (Bell, Boll, Lieb, Long, and Sick) taking turns speaking, with 2-3 speakers visible at any given time.

\begin{table}[!tb]
    \centering
    \small
    \renewcommand{\arraystretch}{1.05}
    \caption{\model{} and state-of-the-art models performance on the different categories of \dataset, grouped by Easy and Hard, using the mean Average Precision (mAP) metric. 
    }
    \begin{tabular}{c*{5}{c}}\hline

        \multirow{2}{*}{\makebox[5em]{\textbf{Model}}} &
        \multicolumn{2}{c}{\makebox[5.5em]{\textbf{Easy}}} & 
        \multicolumn{3}{c}{\makebox[5.5em]{\textbf{Hard}}} \\

        & \textbf{OC} & \textbf{SI} & \textbf{FO} & \textbf{HVN} & \textbf{SS}  \\
        \hline\hline
        
        ASC~\cite{alcazar2020active} & 
         91.2 & 92.3 & 87.1 & 66.8 & 72.2 \\

        MAAS~\cite{alcazar2021maas} & 
         90.7 & 92.6 & 87.0 & 67.0 & 76.5 \\
        
        ASDNet~\cite{kopuklu2021design} & 
         96.5 & 97.4 & 92.1 & 77.4 & 77.8 \\

         TalkNet~\cite{tao2021someone} & 
         95.8 & 97.5 & 93.1 & 81.4 & 77.5 \\

        Light-ASD~\cite{liao2023light} & 
        97.8 &	98.3 &	95.4 &	84.7 & 77.9 \\

        \hline
         \model & 
         \textbf{97.8} & \textbf{98.4} & \textbf{95.9} & \textbf{85.6} & \textbf{82.5} \\
    
        \hline

    \end{tabular}
    \label{table:models-performance-categories}
\end{table}

\begin{table}[!tb]
    \centering
    \small
    \renewcommand{\arraystretch}{1.05}
    \caption{Comparison of \model{}~and state-of-the-art models on the AVA-ActiveSpeaker, grouped by the visual encoder used. Models with $*$ customized the reported backbones.}
    \begin{tabular}{c*{5}{c}}\hline 

      \multirow{2}{*}{\textbf{Model}} &
      \multirow{2}{*}{\textbf{Visual Encoder}} &
       \multirow{2}{*}{\textbf{Par(M)}} &
       \textbf{Body} &
       \multirow{2}{*}{\textbf{mAP}}
       \\

        &  & 
         & \textbf{Data} 
        \\
        \hline\hline
        
        ASC~\cite{alcazar2020active} & 
       RN18 2D & 23.3 & $\times$ & 87.1 \\
        
        MAAS~\cite{alcazar2021maas} & 
       RN18 2D & 21.7 & $\times$ & 88.8 \\
        
        TalkNet~\cite{tao2021someone} & 
        RN18$^*$ 2D-3D & 15.0 & $\times$ & 92.3 \\

        \model &
        RN18$^*$ 2D-3D & 31.6 & $\checkmark$ & \textbf{92.4} \\

        \hline
        
        ASDNet~\cite{kopuklu2021design} & 
         RNx101 3D & 49.7 & $\times$ & 93.5 \\

         EASEE-50~\cite{easee} &
         RN50 3D & 74.7 & $\times$ & 94.1 \\

         Light-ASD~\cite{liao2023light} &
         Conv 2D-1D & 1.0 & $\times$ & 94.1 \\
    
        \hline
        
    \end{tabular}
    \label{table:models-performance-dataset-ava}
\end{table}

\textbf{Models.} The considered models are the ones with state-of-the-art results and publicly available implementations: ASC~\cite{alcazar2020active},  MAAS~\cite{alcazar2021maas}, TalkNet~\cite{tao2021someone}, ASDNet~\cite{kopuklu2021design}, and Light-ASD~\cite{liao2023light}. ASC, MAAS, and ASDNet are trained in a two-step process, while TalkNet and Light-ASD are trained end-to-end. MAAS did not provide its Multi-modal Graph Network setup so we present the results from the available implementation.

\textbf{Evaluation Metrics.} For AVA-ActiveSpeaker and WASD, we use the official ActivityNet evaluation tool~\cite{roth2020ava} that computes mean Average Precision (mAP), while for Columbia we use F1 score. Following the Microsoft COCO Image Captioning Challenge approach, caption generation is evaluated by ROUGE-L~\cite{lin2004rouge}, METEOR~\cite{banerjee2005meteor}, and BLEU-1 to 4~\cite{papineni2002bleu}.

\subsection{\model{}~Performance in \dataset}

To assess the importance of body information for ASD we compare \model{} with the reported results of state-of-the-art models in WASD~\cite{wasd}, divided by categories, in Table~\ref{table:models-performance-categories}.

\textbf{Similar Performance in Easy.}
The inclusion of body information in \model{} culminates in state-of-the-art results across all categories, obtaining slightly better results in Easy setups. In these scenarios, the reliability of face access and sound quality, with minor degradation (OC and SI), is enough to warrant a good performance from state-of-the-art models. As such, the complement of body information from \model{} does not translate into a significant improvement over state-of-the-art models, given the cooperative settings of the Easy group for ASD.

\textbf{Improvement in Hard.}
The major difference of \model{} relative to other models is in Hard categories, with degraded audio and face image quality. In particular, audio degraded categories (HVN and SS) are the ones where \model{} obtains the biggest improvement over state-of-the-art, which is linked to body information access from \model, diminishing the dependence of audio cues for ASD. Regarding scenarios without reliable access to mouth movement (FO), conjunction of body cues with audio information is a more reliable approach for ASD, translated by the increased performance.

\subsection{\model{}~Performance in Other Datasets}

\begin{figure}[!tb]
    \centering
    \includegraphics[width=0.40\textwidth]{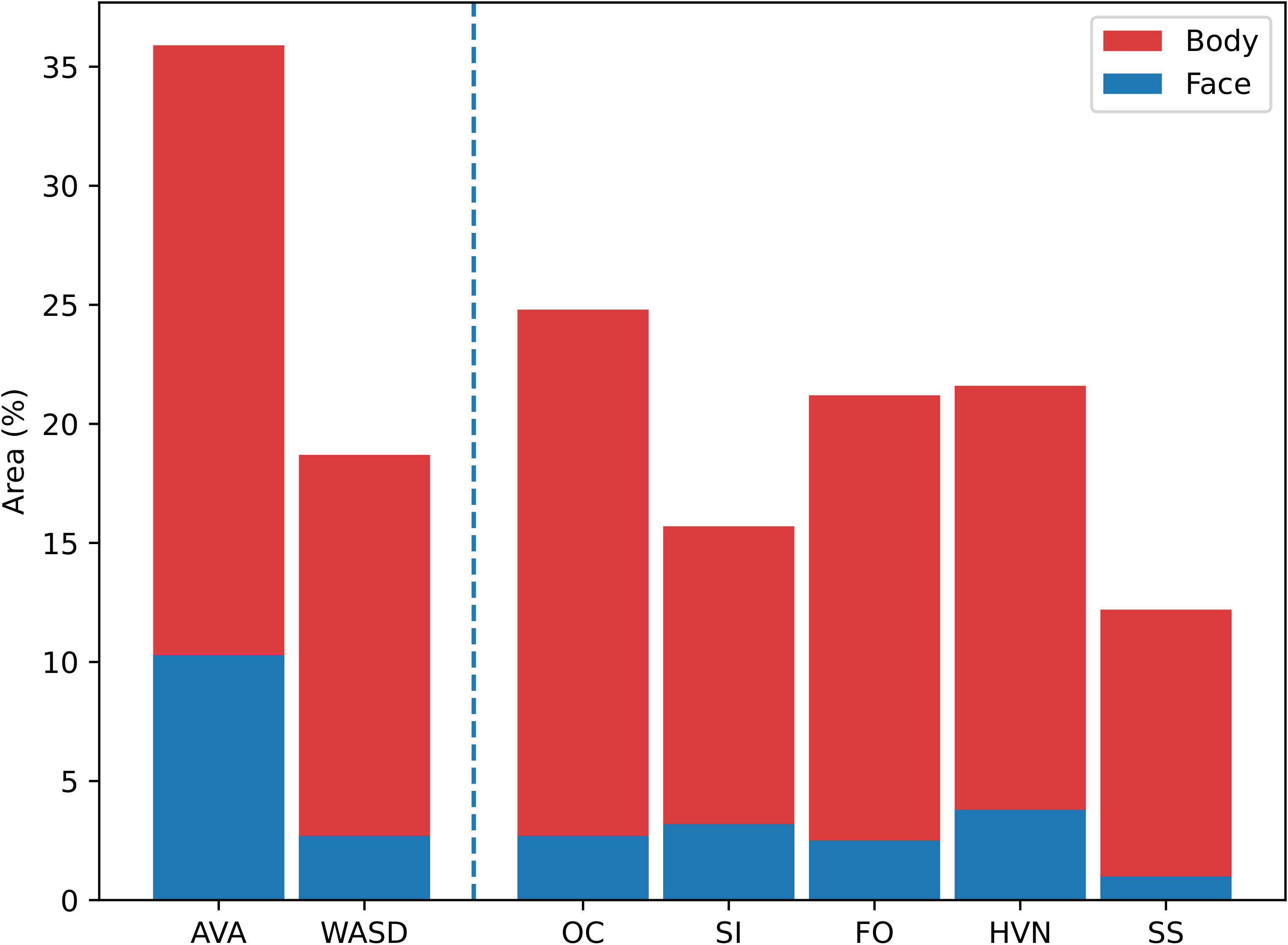}
    \caption{Comparison of face and body area, relative to image dimension, in percentage. 
    AVA-ActiveSpeaker contains data with subjects closer to the camera, expressed by higher face and body percentage, relative to WASD and any of its categories. Surveillance Settings (SS) is the category with further distance of subjects from camera.}
    \label{fig:face_body_area}
\end{figure}

Although we show that inclusion of body information is important for ASD in WASD, in particular for categories with more degraded audio and face data, we also assess its importance in other setups using AVA-ActiveSpeaker and Columbia in Tables~\ref{table:models-performance-dataset-ava} and~\ref{table:columbia}, respectively. 

\textbf{BIAS in AVA-ActiveSpeaker.} We compare \model{} with the other state-of-the-art models, by grouping them into sets, based on the backbone used. The results show that body information inclusion contributes to state-of-the-art performance in ResNet18 grouping, but it is slightly worse than superior or custom backbones, which suggests that body data can be used but it is not as relevant in scenarios where face access and audio quality are more reliable~\cite{wasd}. Figure~\ref{fig:face_body_area} shows the face/body area proportion in AVA-ActiveSpeaker and WASD, grouped by categories, corroborating the reduced importance of body in AVA-ActiveSpeaker data. In this context, the proximity to camera and reliable face access makes body information less contributing to improve ASD, since most of the information usable already exists in the face data. As such, BIAS approach to include body does not translate to a state-of-the-art performance in such settings.

\begin{table}[!tb]
    \centering
    \small
    \renewcommand{\arraystretch}{1.05}
    \caption{Comparison of F1-Score (\%) on the Columbia dataset.}
    \begin{tabular}{cccccc|c}\hline

        \multirow{2}{*}{\makebox[5em]{\textbf{Model}}} &
        \multicolumn{6}{c}{\makebox[5.5em]{\textbf{Speaker}}} \\

        & \textbf{Bell} & \textbf{Boll} & \textbf{Lieb} & \textbf{Long} & \textbf{Sick} & \textbf{Avg} \\
        \hline\hline
        
        TalkNet~\cite{tao2021someone} & 
        43.6 & 66.6 & 68.7 & 43.8 & 58.1 & 56.2 \\

        LoCoNet~\cite{wang2023loconet} & 
        54.0 & 49.1 & 80.2 & 80.4 & 76.8 & 68.1 \\

        Light-ASD~\cite{liao2023light} & 
        82.7 & \textbf{75.7} & 87.0 & 74.5 & 85.4 & 81.1 \\

        \hline

        BIAS & 
        \textbf{89.3} & 75.4 & \textbf{92.1} & \textbf{88.8} & \textbf{88.6} & \textbf{86.8} \\
        \hline

    \end{tabular}
    \label{table:columbia}
\end{table}

\begin{table}[!tb]
    \centering
    \small
    \renewcommand{\arraystretch}{1.05}
    \caption{Ablation studies on the effect of SE for feature combination and the contribution of different features towards BIAS performance (mAP) in AVA-ActiveSpeaker (AVA) and WASD.}
    \begin{tabular}{c*{14}{c}}\hline

        \multicolumn{4}{c}{\makebox[3em]{\textbf{Settings}}} & 
        \multirow{2}{*}{\makebox[3em]{\textbf{AVA}}} &
        \multirow{2}{*}{\makebox[3em]{\textbf{WASD}}} \\

        \textbf{SE} & \textbf{Audio}  & \textbf{Face} & \textbf{Body} & \\
        \hline\hline
        
         $\checkmark$ & 
         $\times$ & $\times$ & $\checkmark$ &
         62.7 & 84.6 \\

         $\checkmark$ & 
         $\times$ & $\checkmark$ & $\times$ &
         79.9 & 90.2 \\

        \hline

         $\checkmark$ & 
         $\checkmark$ & $\times$ & $\checkmark$ &
         75.3 & 86.2 \\

         $\checkmark$ & 
         $\times$ & $\checkmark$ & $\checkmark$ &
         81.2 & 92.1 \\

         $\checkmark$ & 
         $\checkmark$ & $\checkmark$ & $\times$ &
         91.9 & 92.2 \\

         \hline
        $\times$ & 
         $\checkmark$ & $\checkmark$ & $\checkmark$ &
         91.3 & 91.8 \\

         $\checkmark$ & 
         $\checkmark$ & $\checkmark$ & $\checkmark$ &
         92.4 & 94.1 \\

        \hline
 
    \end{tabular}
    \label{table:ablation}
\end{table}

\textbf{Robustness of BIAS in Columbia.} We also assess the performance of BIAS in Columbia, following the methodology of Light-ASD~\cite{liao2023light} where models are trained in AVA-ActiveSpeaker, without any additional fine-tuning, and compare with the results reported on Light-ASD, in Table~\ref{table:columbia}. In this more challenging setting, BIAS approach to combine body with face and audio information leads to a state-of-the-art performance and highlights its robustness to perform ASD in cross-domain settings. Given the results, \model{} is a resilient state-of-the-art ASD model, applicable in scenarios with varying data quality.

\subsection{Ablation Studies}

\textbf{SE Feature Combination.} To complement ASD feature importance for \model{}, we also assess the effect of SE for feature combination and using different features towards BIAS performance in AVA-ActiveSpeaker and WASD, in Table~\ref{table:ablation}. The results show that: 1) When using only one visual feature, face is more relevant than body for both datasets; 2) For two feature combination, audio with face is the approach with better results, particularly for AVA; and 3) The aggregated feature combination is better for both datasets, with increased performance in WASD due to the importance of body information in its most challenging categories; and 4) Regarding the effect of SE, its inclusion translates into improved results for both datasets, with a bigger improvement on WASD, which is linked to its more challenging data where adequate feature selection is of utmost importance (\textit{i.e.}, careful selection of face \textit{vs.} body features is more relevant when face may not be reliably accessed).

\begin{table}[!tb]
    \centering
    \small
    \renewcommand{\arraystretch}{1.05}
    \caption{\model{} performance using different backbones in AVA-ActiveSpeaker (AVA), while maintaining the remaining of the architecture.}
    \begin{tabular}{c*{2}{c}}\hline
        
       \textbf{Visual Backbone} & \textbf{Par(M)} & \textbf{AVA} \\
        \hline\hline

        ResNet18 & 31.6 & 92.4 \\

        ResNet50 & 55.1 & 92.2 \\

        ResNet101 & 91.3 & 92.4 \\
         
        \hline
    \end{tabular}
    \label{table:backbone-variance}
\end{table}

\begin{table}[!tb]
    \centering
    \small
    \renewcommand{\arraystretch}{1.05}
    \caption{\model{} performance (mAP) in AVA-ActiveSpeaker (AVA) and WASD, with varying body inputs and as the only model input for ASD.}
    \begin{tabular}{c*{2}{c}}\hline
        \textbf{Model Input} & \textbf{AVA} & \textbf{WASD} \\
        \hline\hline

        Body w/ Face Region  & 62.7 & 84.6 \\

        Body w/o Face Region & 44.8 & 76.0 \\
     
        \hline
    \end{tabular}
    \label{table:body-vs-face}
\end{table}

\begin{figure}[!tb]
    \centering
    \includegraphics[width=0.45\textwidth]{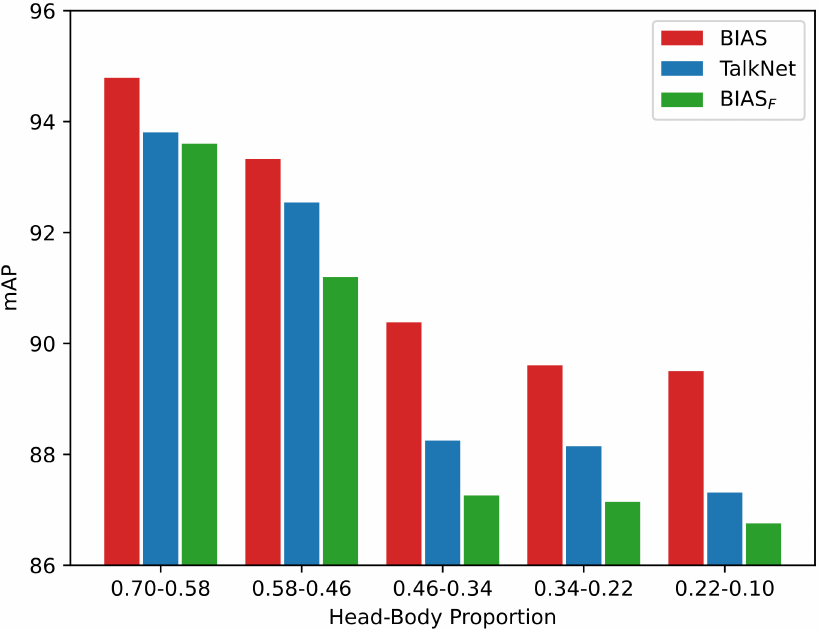}
    \caption{Performance of BIAS, TalkNet, and BIAS$_{F}$ relative to Head-Body Proportion (HBP) in WASD, across 5 equidistant intervals based on minimum (0.1) and maximum (0.7) HBP. BIAS$_{F}$ refers to BIAS with only face as visual input.}
    \label{fig:head-body-mAP}
\end{figure}

\textbf{Backbone Variance.} We explore the influence of having bigger backbones for visual feature extraction in Table~\ref{table:backbone-variance}, which shows varying backbones do not influence BIAS performance. With ASDNet and EASEE, we see that their higher extraction power is also accompanied with additional computationally heavy components to assess actors' relation (\textit{e.g.}, GCN), and customized lightweight models~\cite{liao2023light} also achieve state-of-the-art performance which confirms that simply having higher extraction power does not necessarily translate into better performance. This also occurs in BIAS since the combination of face and body features is not directly influenced by varying the backbone given that the combination is done after feature extraction.

\begin{figure*}[!tb]
    \centering
    \includegraphics[width=0.99\textwidth]{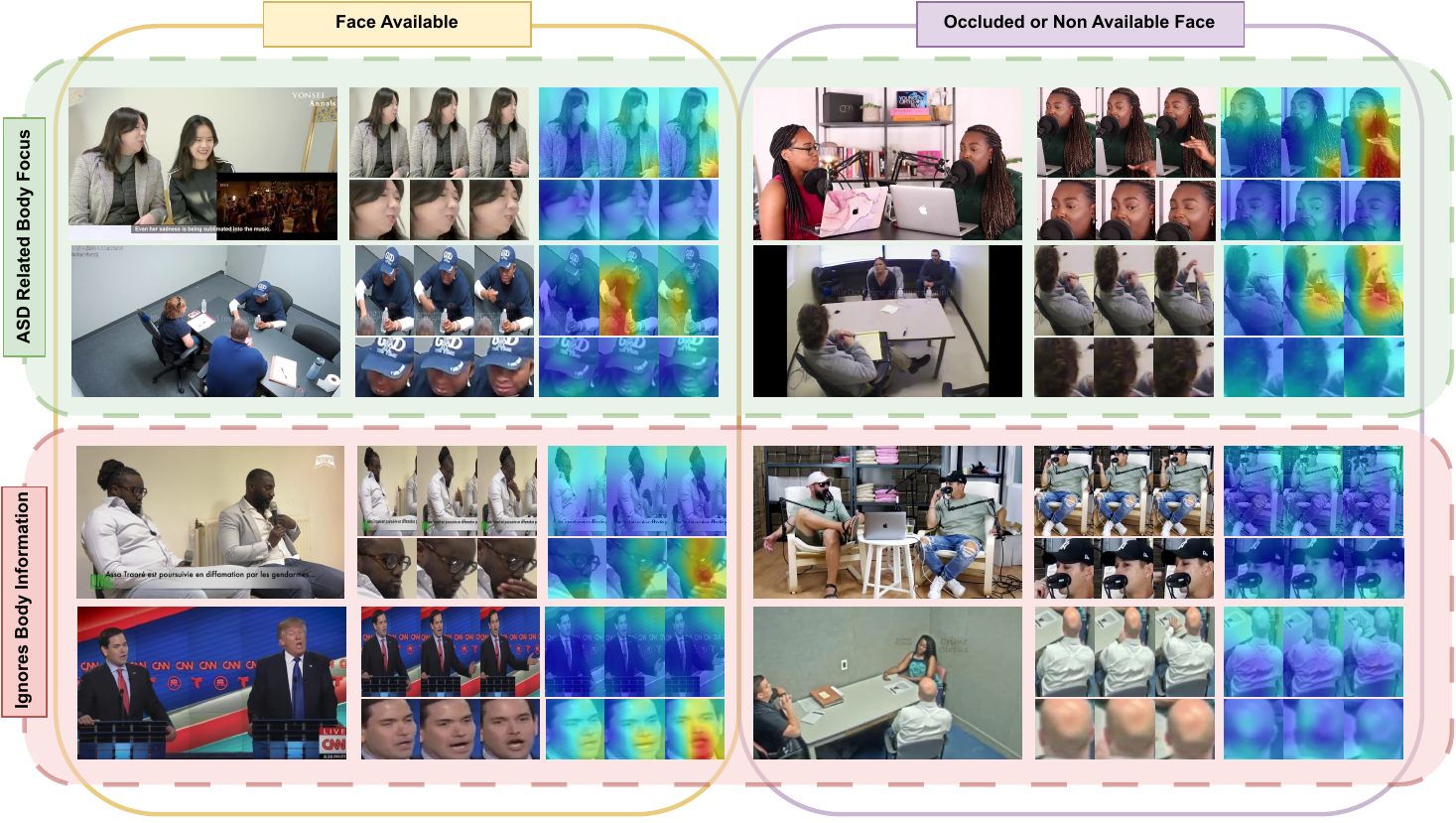}
    \caption{Context, body and face crops, and respective \model{}~attention heatmaps of various scenarios. We group the examples into 4 overlapping sets: 1) face available; 2) occluded or non available face; 3) focus on ASD-body movements, and 4) ignoring body information.}
    \label{fig:body_heatmap}
\end{figure*}

\begin{figure*}[!tb]
    \centering
    \includegraphics[width=\textwidth]{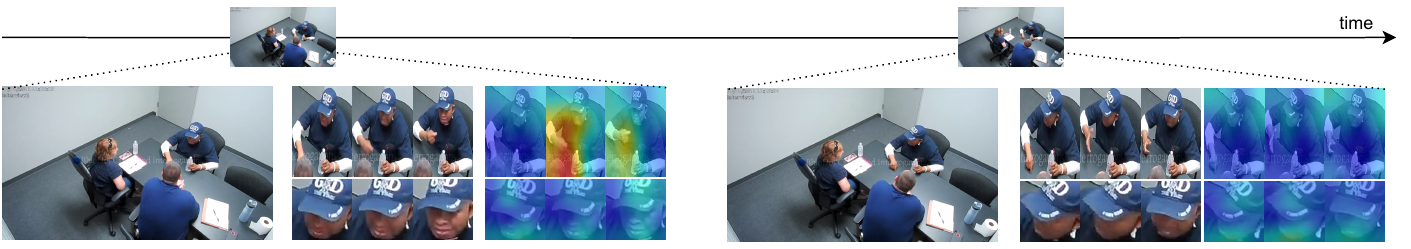}
    \caption{Variation of body movement attention in different time lapses of the same context. Although sudden hand movements related to speech are considered by \model{}, subtle ones are not perceived as important, within the same challenging set.}
    \label{fig:time_body_imp_variation}
\end{figure*}

\textbf{Body \textit{vs.} Face Influence.} Given the novelty of BIAS to include body for ASD, we explore the importance of body \textit{vs.} face influence, in Table~\ref{table:body-vs-face}, where we ignore facial cues of body data by including black boxes in the face region. 
The results show that, although facial cues are important for ASD, body-related actions also contribute towards BIAS performance. However, for AVA-ActiveSpeaker a significant portion of body information comes from the face region, mainly due to the proximity of subjects to the camera in its data (Figure~\ref{fig:face_body_area}), which further supports the inability of BIAS to achieve a state-of-the-art performance in such settings. Our approach in this experiment was to deliberately omit information, which can be seen as an adversarial attack, thus decreasing the model performance~\cite{hingun2023reap, costa2024how}. As such, we also assess the importance of face \textit{vs.} body with varying Head-Body Proportion (HBP), in Figure~\ref{fig:head-body-mAP}, for WASD. 
The results show that when body is significantly predominant relative to face (\textit{i.e.}, low HBP, meaning that face is small), the model that uses body information (BIAS) is superior to others that only use face (BIAS$_{F}$ and TalkNet), and this discrepancy is bigger the lower the HBP is. This shows that the performance discrepancy is independent of the model and is mainly due to the difference of data inputs (\textit{i.e.}, it is not facial cues that justify the differences but rather the information present in body data). As such, even if body information also contains facial cues, body is the most relevant feature when face is small and/or not easily accessed.

\begin{figure*}[!tb]
    \centering
    \subfloat[Audio]{\label{fig:radar_audio}\includegraphics[width=0.33\textwidth]{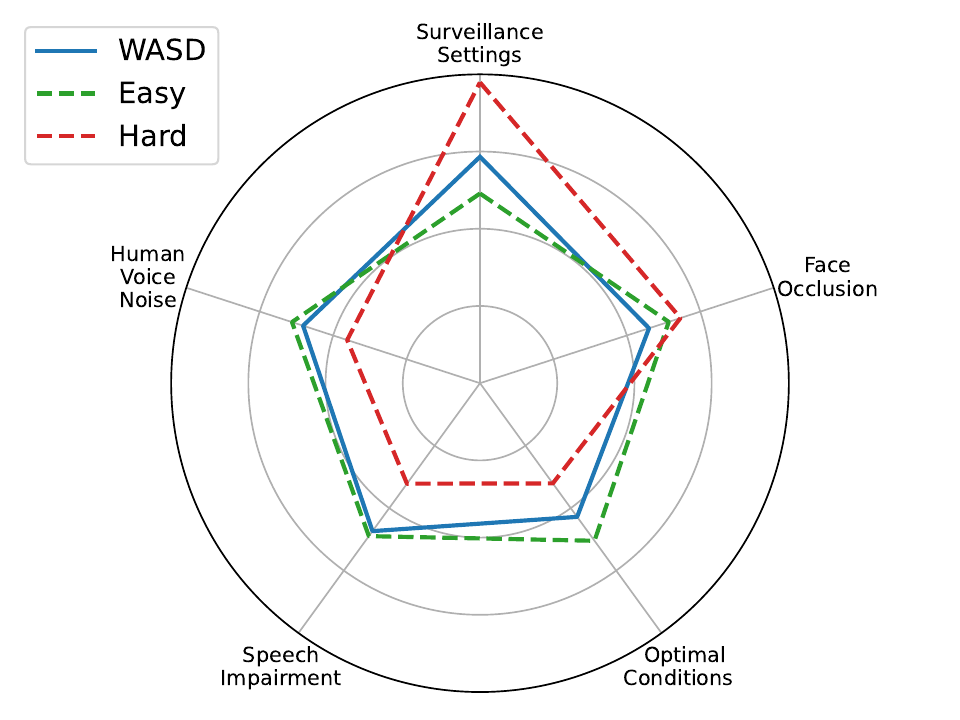}}%
    \subfloat[Body]{\label{fig:radar_body}\includegraphics[width=0.33\textwidth]{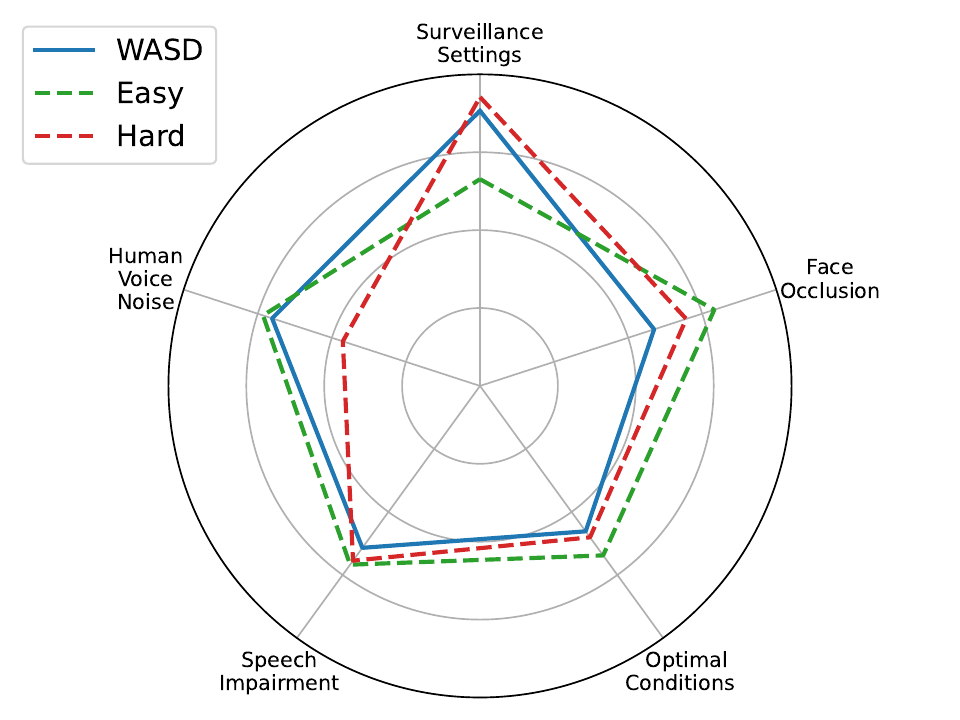}}%
    \subfloat[Face]{\label{fig:radar_face}\includegraphics[width=0.33\textwidth]{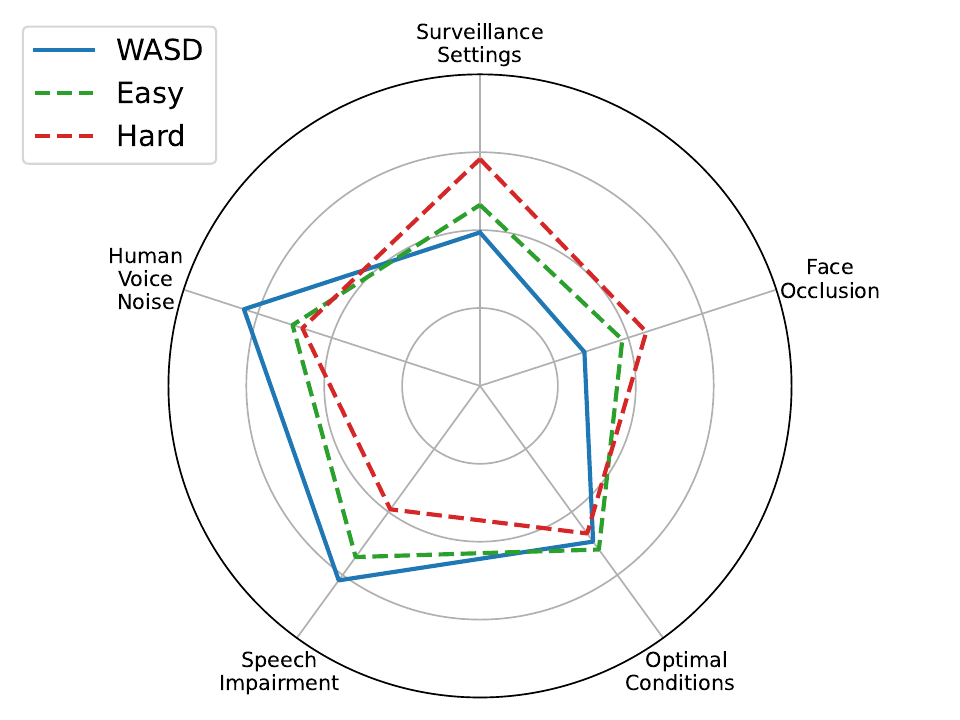}}
    \caption{Relevance of audio, body, and face-based features for ASD on the five different categories of WASD. Hard-\model{} gives higher audio importance in audio impaired categories, perceives body importance similar to WASD-\model{} for the most body-reliant scenario (Surveillance settings) and gives, overall, less face importance to face-dependent contexts, relative to WASD and Easy-\model{}.}
    \label{fig:category_weight_per_feature}
\end{figure*}

\begin{figure*}[!tb]
    \centering
    \subfloat{\includegraphics[width=0.33\textwidth]{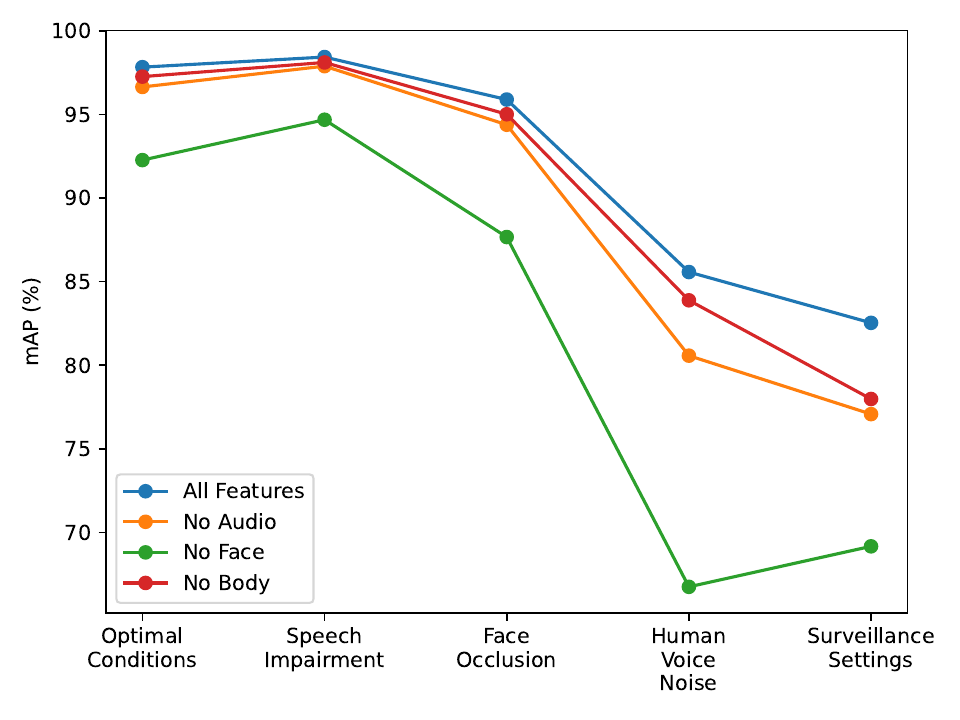}\label{fig:wasd-feature}}%
    \subfloat{\includegraphics[width=0.33\textwidth]{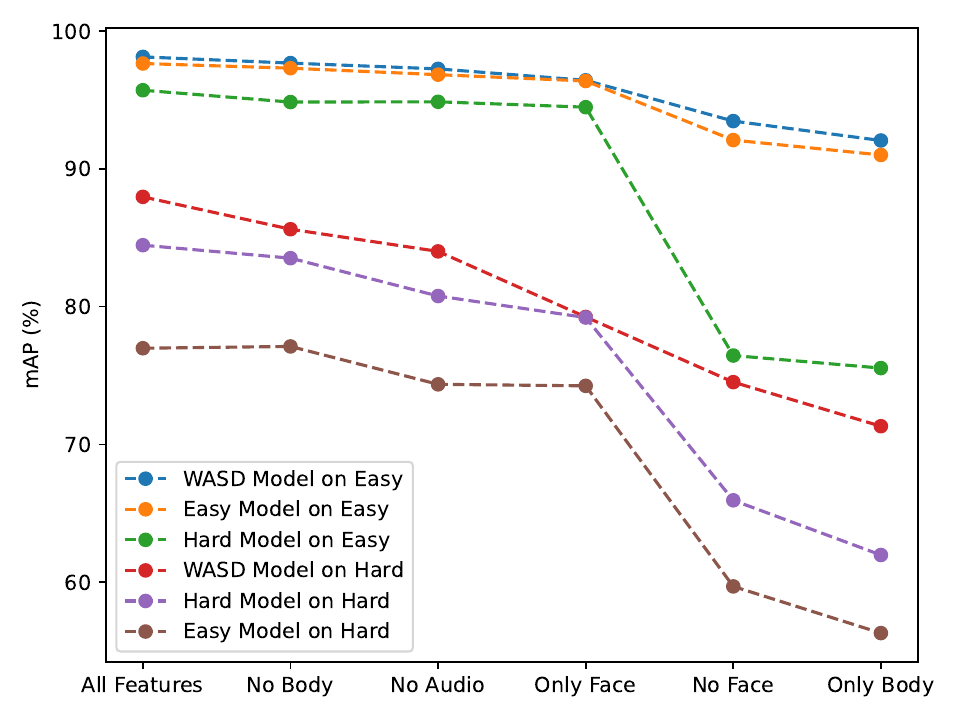}\label{fig:training-effect-feature}}%
    \subfloat{\includegraphics[width=0.33\textwidth]{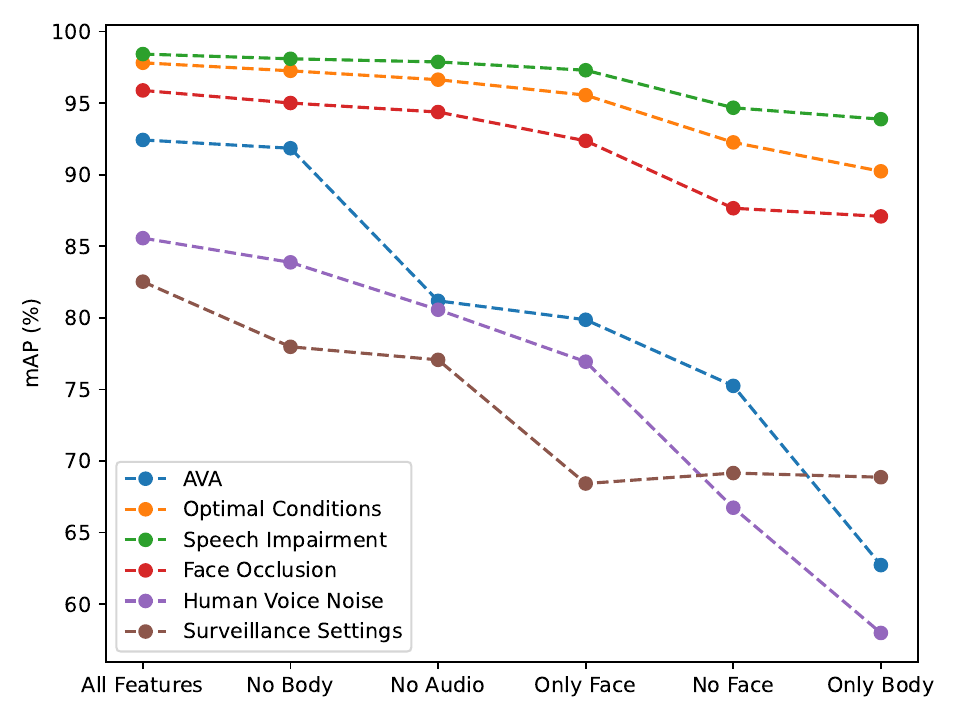}\label{fig:ava-wasd-categories-feature}}
    \caption{Influence of the different features on \model{} performance (mAP): \textit{left}) Effect of audio, face and body absence in WASD categories; \textit{middle}) Feature variation influence relative to different training and testing sets (Easy, Hard, and WASD); and \textit{right}) Feature effect on AVA-ActiveSpeaker (AVA) and WASD categories.}
    \label{fig:mAP_category_model_variation}
\end{figure*}

\section{\model{} Interpretability}
\label{sec:bias-interpretability}

\subsection{Squeeze-and-Excitation Visual Interpretability}

To assess the importance of body for ASD, we create attention heatmaps using SE channels, following the methodology described in Section~\ref{sec:interpretability}. These heatmaps highlight the region where BIAS is focusing on when predicting the active speaker of a given scene. We consider different scenarios to assess the relative importance of face and body via BIAS attention heatmaps, in Figure~\ref{fig:body_heatmap}, where each scenario contains the initial setting (bigger image) and the next three frames for body and face movements (and their respective heatmaps).

\textbf{Body Importance for ASD.} The scenarios in the intersection of \textit{ASD Related Body Focus} with \textit{Face Available} show that BIAS gives higher attention to body when the scenario contains challenging features, even if face is available or partially accessible. In particular, with audio quality subpar (video playing) or limited face availability (hat), body movement importance increases for ASD. When considering the intersection of \textit{ASD Related Body Focus} with \textit{Occluded or Non Available Face}, we are limiting the assess to facial cues which further increases the relative importance of body, translated by BIAS attention to hand movements (in these examples). This shows that body information complement is mainly required in more challenging scenarios, where audio and face are not easily retrieved.  

\textbf{Face over Body.} The intersection of \textit{Ignores Body Information} with \textit{Face Available}, shows examples where BIAS opts to give less relevance to body information (heatmaps of body with little/no coloring). In these cooperative conditions, face is the most relevant/reliable data, leading \model{} to rely exclusively on it and ignore body cues (hand movement in both cases).

\textbf{No Body Focus.} The intersection of \textit{Ignores Body Information} with \textit{Occluded or Non Available Face} shows situations where \model{} can also ignore body information in occluded or non-visible faces. In most cases, BIAS attention to body is mainly linked to pronounced (hand) movement. However, in the scenarios of the intersection, subjects only perform subtle movements when speaking and \model{} response to it is (wrongly) ignoring body information. This shows that BIAS body attention is not entirely linked to speaking activities but more on pronounced/abrupt movements, as exemplified by Figure~\ref{fig:time_body_imp_variation}. With different time frames of the same challenging set, body focus varies in similar scenes (hand movement while talking), with the difference being on the second timeframe having more subtle movement than the first.

\begin{figure}[!tb]
    \centering
    \includegraphics[width=0.5\textwidth]{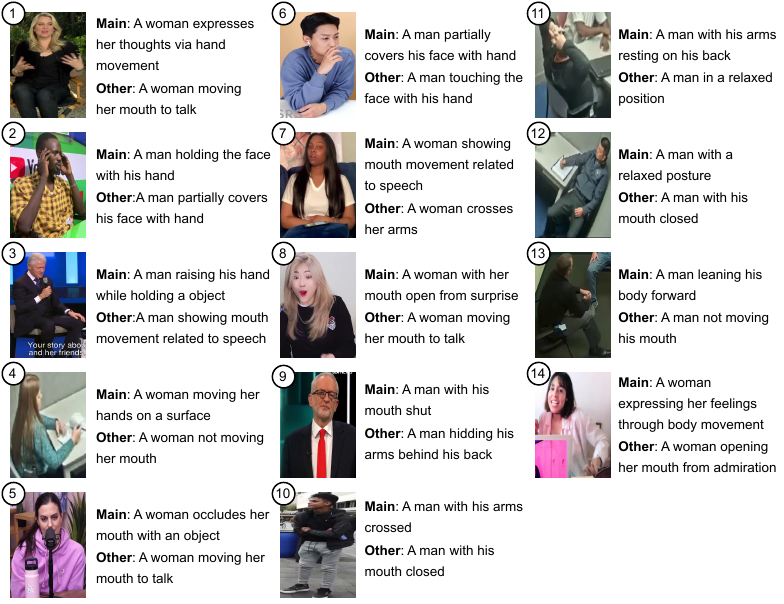}
    \caption{ASD-Text dataset finetuned ViT-GPT2 model predictions in representative images of the 14 considered ASD-related actions of Table~\ref{table:action-gpt}. Each figure number relates to the respective ASD-action number. ViT-GPT2 predictions contain the main caption and a related one (other).}
    \label{fig:gpt_action_example}
\end{figure}

\subsection{Feature Importance Assessment}
\label{subsec:feat_impor_assess}

Although visual explanation provides some reasoning behind \model{} prediction, we also assess the relative importance of features (audio, face, and body data). We compare the effect of training sets on the perceived feature importance, using WASD and Easy/Hard group training, in Figure~\ref{fig:category_weight_per_feature}: WASD-\model{}, Hard-\model{}, and Easy-\model{} refer to \model{} training in WASD, Hard, and Easy data, respectively.

\textbf{Relative Feature Importance.} The main conclusions are: 1) For WASD-BIAS and Easy-BIAS, all the categories have similar audio importance (Figure~\ref{fig:radar_audio}). The most divergent variation is Hard-BIAS, which gives less importance to audio in categories with faces available and higher importance in scenarios with occluded faces; 2) Both WASD-BIAS and Hard-BIAS give higher importance to body in surveillance scenarios relative to other categories (Figure~\ref{fig:radar_body}), given the various challenges in this setting. Furthermore, with Hard-BIAS, body is less important for the category with more close-up faces and less visible body. This trend is not seen in the other models given their different training conditions (\textit{i.e.}, they have access to other categories with close-up faces and less visible body), leading to a well rounded body importance; and 3) Regarding face importance (Figure~\ref{fig:radar_face}), Hard-BIAS is the more evenly distributed model, meaning that face is overall important for ASD regardless of the category. Regarding WASD-BIAS (and Easy-BIAS to some extent), face is more relevant for scenarios with close-up faces and impaired audio, leading to increased face importance relative to the other categories.

\textbf{Performance Influence.} We explore the effect of group training and feature influence in model performance, in Table~\ref{table:train-effect} and Figure~\ref{fig:mAP_category_model_variation}, respectively. Overall, face is the most influential feature for~\model{} in  WASD, and the removal of body information has higher impact in surveillance settings, which highlights the body importance in these conditions (Figure~\ref{fig:wasd-feature}). When considering the impact of different features (audio, face, and body) in different training sets (Figure~\ref{fig:training-effect-feature}), we observe that subsequent removal of available features translates into decreased performance, with training with all data (WASD) translating in more resilient results than training only on Easy or Hard data alone. This and the results in Table~\ref{table:train-effect} highlight the importance of robust training data for improved resilience, particularly to avoid situations where training may lead to a false sense of face reliability (drop of performance from Hard-BIAS on Easy data when not using face). We also extend the analysis of feature influence in AVA and WASD training (Figure~\ref{fig:ava-wasd-categories-feature}). With subsequent removal of features, BIAS underperfoms the most on categories with audio impairment (HVN) and unreliable face access and audio quality (SS). Furthermore, surveillance settings is where body information is of utmost importance, based on the performance stability with using \textit{Only Face}, \textit{No Face} and \textit{Only Body} (\textit{i.e.} good face/body importance balance). Finally, \model{} training in AVA is not reliant on body, and has greater focus on audio and face features, heavily degrading in performance without these features. This shows that AVA data does not promote adequate feature combination, translating in models less resilient to varying data quality (\textit{i.e.}, when audio and/or face quality are affected).

\begin{table}[!tb]
    \centering
    \small
    \renewcommand{\arraystretch}{1.05}
    \caption{Effect of group training, in mAP, on the categories of \dataset. Easy group training contains data of Optimal Conditions and Speech Impairment, while Hard group contains data of the remaining WASD categories.}
    \begin{tabular}{c*{14}{c}}\hline

        \multirow{2}{*}{\makebox[5em]{\textbf{Category}}}  &
        \multicolumn{3}{c}{\makebox[5.5em]{\textbf{Train Set}}} \\

        & \textbf{Easy} & \textbf{Hard}  & \textbf{WASD} \\
        \hline\hline
        
        Optimal Conditions & 
         97.1 & 94.7 & 97.8 \\

         Speech Impairment & 
         98.2 & 96.7 & 98.4 \\
        
         Face Occlusion & 
         90.4 & 93.5 & 95.9 \\

         Human Voice Noise & 
         77.3 & 83.4 & 85.6 \\

         Surveillance Settings & 
         63.2 & 79.4 & 82.5 \\

        \hline

    \end{tabular}
    \label{table:train-effect}
\end{table}

\begin{table}[!tb]
    \centering
    \small
    \renewcommand{\arraystretch}{1.05}
    \caption{ViT-GPT2 performance increase with finetune on ASD-Text dataset. Metric description: RL, ROUGE-L; M, METEOR; B1-4, BLEU1-4.}
    \begin{tabular}{c*{6}{c}}\hline
        Model & \textbf{RL} & \textbf{M} & \textbf{B1} & \textbf{B2} & \textbf{B3} & \textbf{B4} \\
        \hline\hline
        
         ViT-GPT2$_{Base}$ & 
         0.28 & 0.18 & 0.14 & 0.03 & 0.01 & 0 \\

         ViT-GPT2$_{ASD}$ & 
         0.61 & 0.58 & 0.61 & 0.50 & 0.37 & 0.31 \\
         
        \hline
    \end{tabular}
    \label{table:gpt-performance}
\end{table}

\begin{figure}[!tb]
    \centering
    \subfloat{\includegraphics[width=0.25\textwidth]{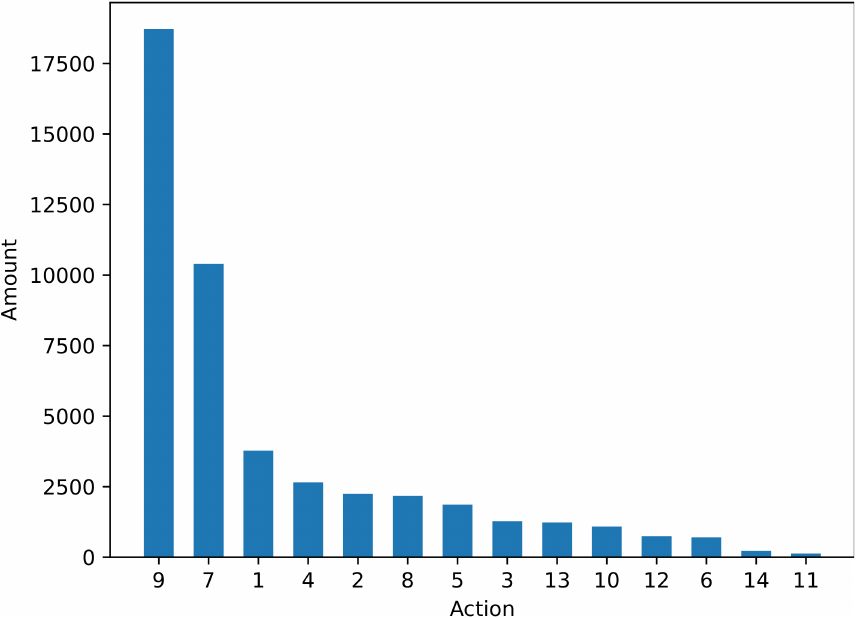}}%
    \subfloat{\includegraphics[width=0.25\textwidth]{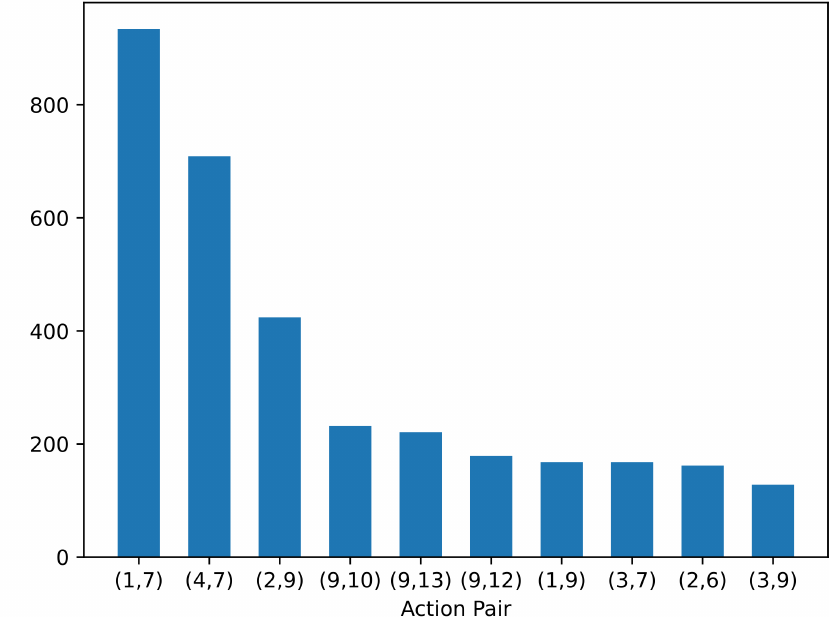}}
    \caption{Distribution of ASD-Text dataset actions (left) and pair of actions more commonly associated (right). Action labels refer to their numbers in Table~\ref{table:action-gpt}.}
    \label{fig:asd-action-distribution}
\end{figure}

\begin{figure*}[!tb]
    \centering
    \includegraphics[width=0.9\textwidth]{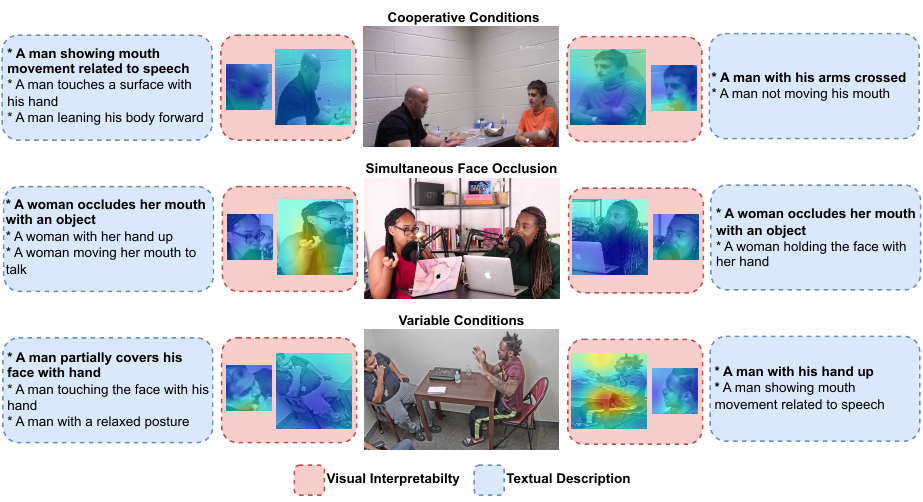}
    \caption{Combination of SE visual interpretability with ViT-GPT2 model subject description in various scenarios with varying ASD challenges. Bold descriptions are the main prediction of ViT-GPT2.}
    \label{fig:visual_textual_interpretability}
\end{figure*}

\subsection{ASD-Text Dataset}
\label{sec:text-description}

To improve the interpretability of ASD-related scenarios, we complement the visual interpretability of attention heatmaps with text scene description, via caption generation of a ViT-GPT2 model. Given the absence of relevant data to train models for captions on ASD-related actions, we start by showing the relevance of the proposed ASD-Text dataset (Section~\ref{sec:asd-textual}), annotation distribution, and its importance towards a full interpretability setup.

\textbf{ASD-Text Dataset Importance.} We assess the performance of a pretrained (Base) and finetuned (ASD) ViT-GPT2 on ASD-Text dataset, on Table~\ref{table:gpt-performance}, where the results support the need for ASD-Text dataset given the improvements of a finetuned ViT-GPT2 in predicting ASD-related captions. Based on the increase of all caption related metrics to reference levels~\cite{wang2020diversity}, finetuned ViT-GPT2 can reliably describe ASD scenes in various challenging scenarios (WASD data). Furthermore, we also display ViT-GPT2 predictions on representative examples of the considered training actions (Table~\ref{table:action-gpt}), in Figure~\ref{fig:gpt_action_example}. We present two captions per image based on the top 5 predictions of ViT-GPT2, with \textit{main} being its first predicted caption and \textit{other} being the first caption not related to the action of the main.

\textbf{Annotation Distribution.} Figure~\ref{fig:asd-action-distribution} shows the overall distribution of ASD-Text dataset actions. The majority of subjects have mouths closed or actively moving, relating to the speaking labels considered (talking \textit{vs.} not talking). Regarding the actions more commonly paired, we have subjects talking with hand movement, either raised or on a surface, illustrating body movement associated with talking. More linked to non-talking subjects we have hands touching faces, without occluding mouth, as the most common pair association. 

\textbf{Full Interpretability Setup.} We can also use ASD-Text dataset to create a complete interpretability setup that combines the visual information of \model{} (attention heatmap)~with the text description of ViT-GPT2, as shown in Figure~\ref{fig:visual_textual_interpretability}. In the three different scenarios, with varying challenges, visual information and text scene description represent the key characteristics for the decision behind accurate ASD.

\section{Conclusion}
\label{sec:conclusion}

In this paper we propose BIAS, a multi-modal approach for Active Speaker Detection (ASD) that singularly considers audio, face, and body-based information, which is state-of-the-art in challenging settings and has competitive results in more cooperative conditions. Furthermore, we propose a novel application of Squeeze-and-Excitation blocks to assess ASD feature importance in different settings and provide visual interpretability, complementing them with text descriptions from a ViT-GPT2 model (finetuned in ASD-Text dataset) for a full interpretability setup. Our work highlights the importance of body inclusion for ASD in unconstrained/challenging conditions and serves as baseline for models to perform in wilder scenarios such as surveillance settings.

\section*{Acknowledgments}

This work was supported in part by the Portuguese FCT/MCTES through National Funds and co-funded by EU funds under Project UIDB/50008/2020; in part by the FCT Doctoral Grant 2020.09847.BD and Grant 2021.04905.BD.






 




\begin{IEEEbiography}[{\includegraphics[width=1in,height=1.25in,clip,keepaspectratio]{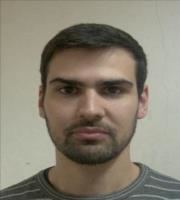}}]{Tiago Roxo} (Member, IEEE) obtained a bachelor's degree in Computer Science and Engineering from Universidade da Beira Interior (UBI) in 2019 and is currently pursuing a Ph.D.'s degree, with a FCT (\textit{Funda\c{c}\~{a}o para a Ciência e a Tecnologia}) scholarship, in the field of Computer Vision and Artificial Intelligence.
\end{IEEEbiography}

\begin{IEEEbiography}[{\includegraphics[width=1in,height=1.25in,clip,keepaspectratio]{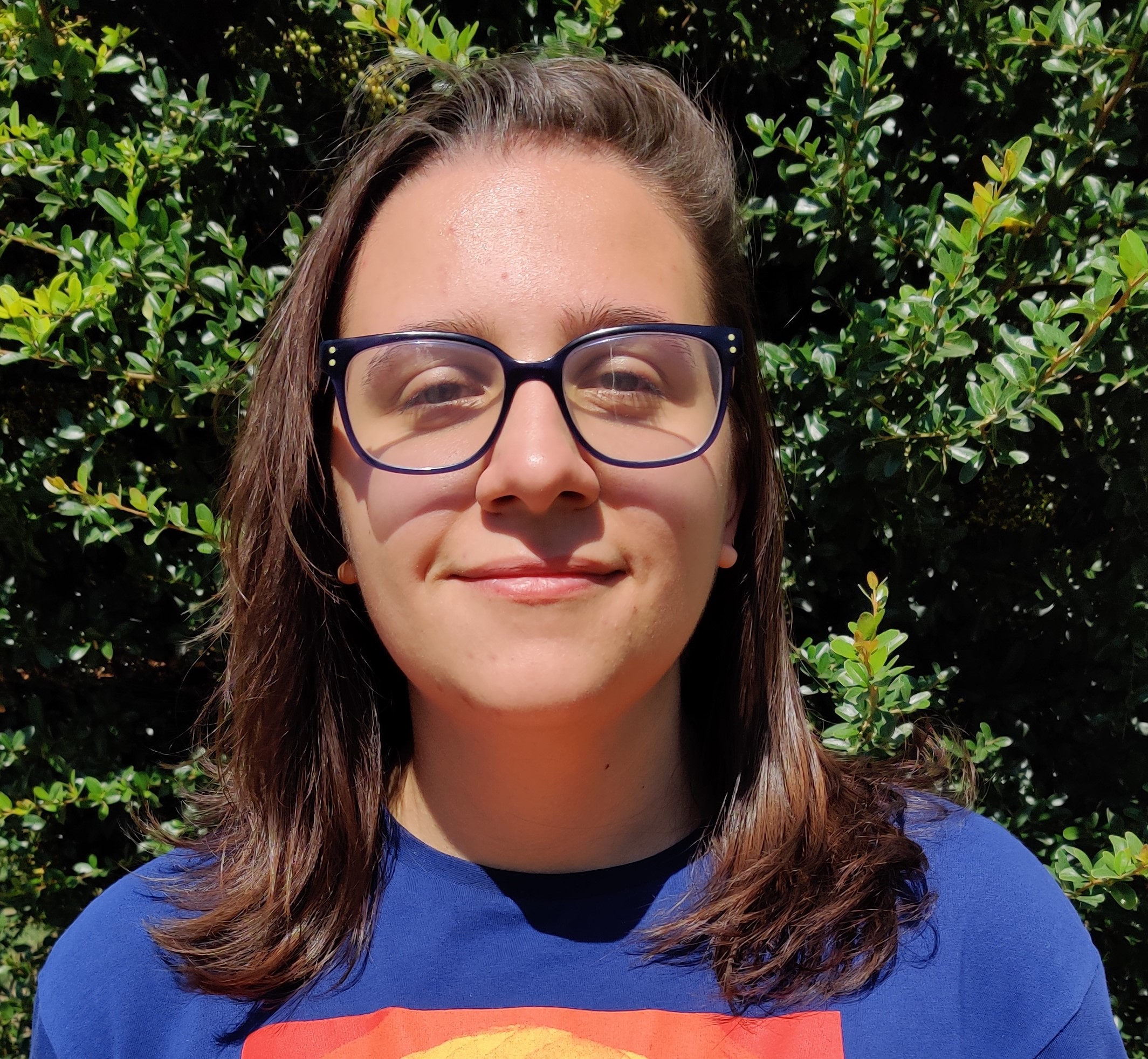}}]{Joana Cabral Costa} obtained her bachelor's and master's degree in Computer Science and Engineering from Universidade da Beira Interior (UBI) in 2019 and 2021, respectively. She is currently pursuing a Ph.D.'s degree, with a FCT (\textit{Funda\c{c}\~{a}o para a Ciência e a Tecnologia}) scholarship, in the field of Computer Vision and Adversarial Attacks.
\end{IEEEbiography}

\begin{IEEEbiography}[{\includegraphics[width=1in,height=1.25in,clip,keepaspectratio]{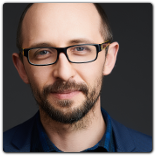}}]{Pedro R. M. In\'{a}cio} is an associate professor of the Department of Computer Science at the University of Beira Interior (UBI), which he joined in 2010. Lectures subjects related with information assurance and (cyber)security, and computer based simulation, to graduate and undergraduate courses, namely to the B.Sc., M.Sc. and Ph.D. programmes in Computer Science and Engineering. He is currently the Pro-Rector for the Digital University and the Data Protection Officer of UBI. Holds a 5-year B.Sc. degree in Mathematics/Computer Science and a Ph.D. degree in Computer Science and Engineering, obtained from UBI, Portugal, in 2005 and 2009 respectively. The Ph.D. work was performed in the enterprise environment of Nokia Siemens Networks Portugal S.A., through a Ph.D. grant from the Portuguese Foundation for Science and Technology.

He is an IEEE senior member, an ACM professional member and a researcher of the Instituto de Telecomunicações (IT). His main research topics are information assurance and security, computer based simulation, and network traffic monitoring, analysis and classification. He has 70+ publications in the form of book chapters and papers in international peer-reviewed books, conferences and journals. He frequently reviews papers for IEEE, Springer, Wiley and Elsevier journals. He is a member of the Technical Program Committees of flagship national and international workshops and conferences, such as ACM SAC, IEEE NCA, IFIPSEC or ARES.
\end{IEEEbiography}

\begin{IEEEbiography}[{\includegraphics[width=1in,height=1.25in,clip,keepaspectratio]{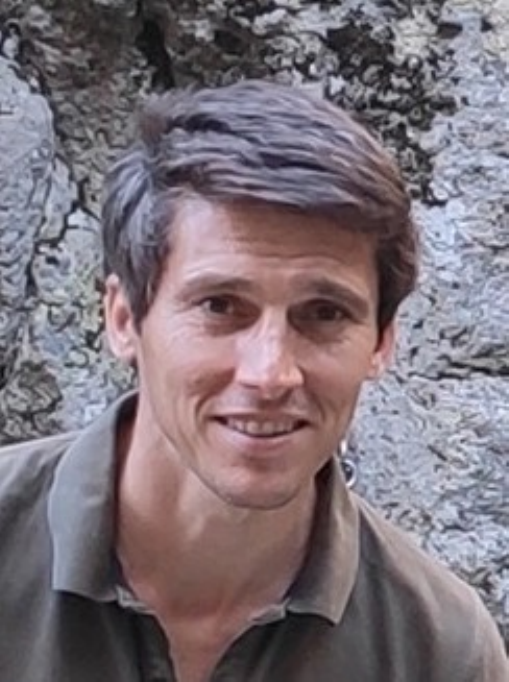}}]{Hugo Proen\c{c}a} (SM'12), B.Sc. (2001), M.Sc. (2004) and Ph.D. (2007) is an Associate Professor in the Department of Computer Science, University of Beira Interior and has been researching mainly about biometrics and visual-surveillance. He was the coordinating editor of the IEEE Biometrics Council Newsletter and the area editor (ocular biometrics) of the IEEE Biometrics Compendium Journal. He is a member of the Editorial Boards of the Image and Vision Computing, IEEE Access and International Journal of Biometrics. Also, he served as Guest Editor of special issues of the Pattern Recognition Letters, Image and Vision Computing and Signal, Image and Video Processing journals. 
\end{IEEEbiography}


\end{document}